\documentclass[a4paper,fleqn]{cas-dc}

\usepackage[authoryear]{natbib}
\usepackage[baseline]{euflag}
\usepackage[switch]{lineno}
 
\def\tsc#1{\csdef{#1}{\textsc{\lowercase{#1}}\xspace}}
\tsc{WGM}
\tsc{QE}
\newcommand{\cready}[1]{\textcolor{black}{{#1\\}}}

\begin{document}
\let\WriteBookmarks\relax
\def\floatpagepagefraction{1}
\def\textpagefraction{.001}

\shorttitle{Shape and Style GAN-based Multispectral Data Augmentation}    

\shortauthors{Fawakherji \textit{et al.}}  

\title [mode = title]{Shape and Style GAN-based Multispectral Data Augmentation for Crop/Weed Segmentation in Precision Farming}



%

\author[1]{Mulham Fawakherji}[orcid=0000-0003-0060-2637]
\cormark[1]
\ead{mfawakherji@ncat.edu}


\affiliation[1]{organization={Department of Built Environment, College of Science and Technology North Carolina A\&T University},
            addressline={1601 E Market St}, 
            city={Greensboro},
            postcode={27411}, 
            country={US}}

\affiliation[2]{organization={Department of Computer, Control, and Management Engineering Antonio Ruberti, Sapienza University of Rome},
            addressline={Via Ariosto 25}, 
            city={Roma},
            postcode={00185}, 
            country={Italy}}


\affiliation[3]{organization={Faculty of Political Science and Sociopsychological Dynamics, UNINT University},
            addressline={Via Cristoforo Colombo, 200}, 
            city={Roma},
            postcode={00147}, 
            country={Italy}}

\author[2]{Vincenzo Suriani}[orcid=0000-0003-1199-8358]
\ead{suriani@diag.uniroma1.it}
\author[2]{Daniele Nardi}[orcid=0000-0001-6606-200X]
\author[3]{Domenico Daniele Bloisi}[orcid=0000-0003-0339-8651]
\ead{domenico.bloisi@unint.eu}
\ead{nardi@diag.uniroma1.it}

\cortext[1]{Corresponding author}



\begin{abstract}
The use of deep learning methods for precision farming is gaining increasing interest. However, collecting training data in this application field is particularly challenging and costly due to the need of acquiring information during the different growing stages of the cultivation of interest. In this paper, we present a method for data augmentation that uses two GANs to create artificial images to augment the training data.
To obtain a higher image quality, instead of re-creating the entire scene, we take original images and replace only the patches containing objects of interest with artificial ones containing new objects with different shapes and styles. In doing this, we take into account both the foreground (i.e., crop samples) and the background (i.e., the soil) of the patches. Quantitative experiments, conducted on publicly available datasets, demonstrate the effectiveness of the proposed approach. The source code and data discussed in this work are available as open source.
\end{abstract}



\begin{keywords}
 Crop protection \sep Crop/weed detection \sep Data augmentation \sep Multispectral image segmentation \sep Precision agriculture 
\end{keywords}

\maketitle

\section{Introduction}

Modern agriculture is undergoing a transformative revolution, driven by the integration of artificial intelligence (AI) technologies into farming practices. Among these AI-powered advancements, \textit{Precision Agriculture} has emerged as a promising approach to optimize resource utilization and to enhance crop yield. It aims at improving crop yields thus increasing productivity. New technologies can play a relevant role in this field, leading to more sustainable agricultural production and better management of natural resources. This paradigm shift relies on cutting-edge technologies such as deep learning algorithms for the accurate detection and management of crops and weeds in cultivated fields. 

However, deep learning models require a large amount of training examples to work properly. Collecting a large training dataset involves a considerable time effort, especially in the case of pixel-wise labeling, where each pixel in each image has to be labeled individually.
In addition to the difficulty of acquiring a large amount of data, it is important also to take into account the class distribution, which is usually imbalanced, meaning that one specific class has a higher number of instances than others \citep{Wang6170916}.
In the case of class imbalance, the classifier could be less accurate when searching for the decision boundaries. 

In some scenarios, unbalanced datasets are more frequent and common data augmentation techniques are not suitable, due to possible color and shape variation over time of the objects of interest and the presence of varying light conditions. \textit{Precision Agriculture} is one of those scenarios.

 Classifying crops and weeds for targeted interventions on a single plant is a crucial point for applying precision agriculture technologies. However, collecting samples for this kind of task is still challenging due to the variability of the environmental conditions and the large variety of crops and weeds. In fact, samples should be acquired across different weather conditions and growth stages (including variations in shapes, size, and colors).

\begin{figure}[!t]
    \centering
    \includegraphics[width=0.9\columnwidth]{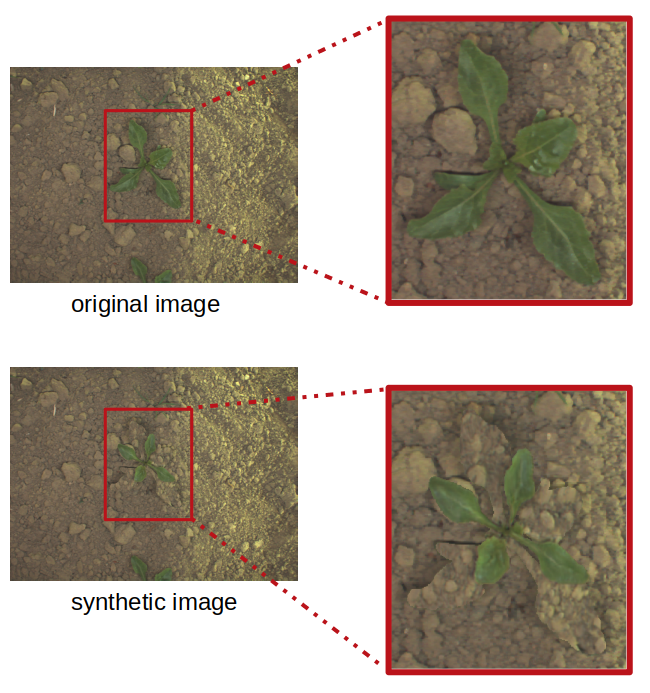}
    \caption{Synthetic image generation. The real plant in the patch highlighted by a red box in the original image is replaced by a synthetic plant with a different shape and style.
    }
    \label{fig:example}
\end{figure}



In this paper, we describe a solution for the class imbalance problem
in crop/weed segmentation. Our idea is to generate only the objects that belong to the minority classes that are relevant for semantic segmentation purposes (see Fig.~\ref{fig:example}). To do so, we first generate new crop shapes using a Deep Convolutional Generative Adversarial Network (DCGAN). Then, we use a conditional Generative Adversarial Network (cGAN) to generate synthetic style samples of the target object. Finally, we replace the real target object (minority class) with a synthetically generated one, keeping the rest of the image (majority class) as it is (i.e., without modifications).

The contribution of this work is three-fold.
\begin{enumerate}
    \item We propose an architecture composed of a DCGAN and a cGAN to achieve shape\&style data augmentation.
    \item We provide a solution for keeping the verisimilitude of the synthetic data high by conditioning both the shape and the style of the generated images.
    \item Our approach is designed to work with multispectral images, which are very useful in precision agriculture applications.
\end{enumerate}

Moreover, the source code and the data generated by our method are made publicly available at \url{www.sites.google.com/diag.uniroma1.it/shapestyle}.

The remainder of the paper is organized as follows.
Section \ref{sec:relwork} presents a brief overview of related work.
Our approach is detailed in Section \ref{sec:method}, while 
experimental results are shown in Section \ref{sec:results}.
Section \ref{sec:conclusions} provides the conclusions.

\section{Related Work}
\label{sec:relwork}
\cready{The crop/weed segmentation problem has garnered significant attention in recent years, prompting extensive research efforts \citep{LU2022107208}. In the early stage, many researchers used traditional machine learning methods that utilize hand-crafted features to identify a set of distinguishing features that will be useful in discriminating between plant classes. }
For example, \citep{NGUYENTHANHLE2019116} suggests using multi-feature algorithms based on shape and color features to detect the weed in a soybean field.
\citep{zhang2019selfattention} analyze different color spaces such as  RGB, HSV, and HIS trying to extract common features for different types of weeds at the pea seedling stage. 

Other approaches aim at improving the generalization capability of traditional machine learning methods by using images captured within the different wavelength ranges across the electromagnetic spectrum like multi-spectral images. For example,
\citep{lottesJFR2016} propose to use a multi-spectral camera to detect weeds in sugar beet fields. Their method starts with detecting vegetation, then an object-based features extraction is implemented, followed by a random forest classification, and finally, they apply a smoothing post-process through a Markov random field.

Hand-crafted feature-based (and derived) methods suffer the dependency from the choice of the features and this can limit the robustness of the system. 
A solution to increase the robustness and the generalization capabilities of these systems comes from the use of Neural Network methods.
For example,
\cite{potena2016} apply a cascade of two Convolutional Neural Networks (CNNs) to the crop/weed classification task, while
\cite{mccool2017mixtures} propose a three-stage approach with the use of model compression techniques and mixtures of models. 

If CNNs are very common and useful in classification, Semantic Segmentation Deep Neural Networks are convenient for achieving segmentation.
One of the most commonly adopted approaches for crop/weed segmentation is SegNet \citep{Segnet}. For example,
\cite{dicicco2017} train SegNet with real and synthetic images achieving good segmentation performance. Also
\cite{weednet} use SegNet for dense semantic weed classification on multispectral images.

\cite{miliotoicra2018} augment the RGB input image with task-relevant background knowledge, allowing to increase the generalization capability of the network.
Relying on the same mechanism, we present in \cite{FawakherjiYBPN19} a pipeline with two CNNs, one for pixel-wise segmentation and the other one for classification, which exploits data coming from different contexts to achieve a good generalization with respect to different types of crop. 

Although both CNNs and Semantic Segmentation Networks proved to be useful technologies, their applicability is limited by the need for a large quantity of data in the training phase.
In the field of precision farming, the collection of large annotated data requires a notable effort in terms of time. First of all, data have to be collected across the weed growth stages and under different weather conditions. Then, once the data are available, the labeling process can be very time-consuming, especially when labeling is pixel-wise. To tackle this problem, it is possible to shrink an unlabeled dataset by preserving only the most informative images while keeping a sufficient segmentation performance \citep{potena2016}. Also, a graphic engine can be used to generate synthetic farming scenes, which contain natively the corresponding ground truth data \citep{dicicco2017}.
\cite{miliotoicra2018} propose a CNN that requires a limited amount of data to generalize to unseen environments with high segmentation accuracy.

New approaches have taken advantage of GANs. For example, \cite{valerio2017arigan} propose a GAN capable of generating \textit{Arabidopsis} plants, allowing to condition the generation by the desired number of leaves for the synthetically created plants. Another application of the GANs is presented by
\cite{MADSEN2019147}.\cready{ They generate synthetic image samples of plant seedlings to compensate for a lack of training data. In particular, nine distinct species of plants are generated, improving the overall accuracy. }
\cready{In \cite{espejo2021combining}, synthetic RGB images of individual tomato and black night-shade plants are generated for improving classification using a GAN. In \cite{khan2021novel}, artificial data generated from UAV images by means of Semisupervised GANs is used for supporting crop/weed species identification at an early stage.}

\cready{In \citep{KIM2022107146}, authors propose a multi-task semantic segmentation-convolutional neural network for detecting crops and weeds (MTS-CNN) using one-stage training. More recently, \citep{divyanth2022image} aims to curtail the effort needed to prepare very large image datasets by creating artificial images of maize and four common weeds through conditional GAN (cGANs). The style of leaves is preserved in  \citep{xu2022style}, where images in the source domain are translated into the target domain. In contrast, the variations unrelated to the domain are maintained to augment the dataset. As a difference from other existing approaches, in our method, the foreground and the background are generated altogether and the new artificial samples are generated using jointly RGB and NIR data. A comparison with similar state-of-the-art approaches is shown in Table \ref{tab:comparison}. }

\begin{table*}[]
\caption{Comparison across recent approaches using GANs for synthetic data generation in precision agriculture}
\begin{tabular}{cccc}
\hline
Approach                                                                                                                                                  & \begin{tabular}[c]{@{}c@{}}Crop \\ Production\end{tabular}     & \begin{tabular}[c]{@{}c@{}}Image Synthesis\\ Technique\end{tabular}     & Results/Conclusion                                                                                                                                                  \\ \hline
\multicolumn{4}{c}{\citep{espejo2021combining}}                                                                                                                                                                                                                                                                                                                                                                                                           \\ \hline
\begin{tabular}[c]{@{}c@{}}Synthetic RGB images of individual\\ tomato and black night-shade plants\\ generated for improving classification\end{tabular} & Tomato                                                         & \begin{tabular}[c]{@{}c@{}}Conventional \\ GANs\end{tabular}            & \begin{tabular}[c]{@{}c@{}}F1-score of 0.86 obtained with \\ GAN-based augmentation, \\ compared to 0.84\\ without the artificial dataset.\end{tabular}             \\ \hline
\multicolumn{4}{c}{\cite{FAWAKHERJI2021103861}}                                                                                                                                                                                                                                                                                                                                                                                                           \\ \hline
\begin{tabular}[c]{@{}c@{}}Generation of multi-spectral images \\ of agricultural fields for semantic\\ segmentation of crop/weeds\end{tabular}           & Sugarbeet                                                      & \begin{tabular}[c]{@{}c@{}}Conditional \\ GAN\\ (cGAN)\end{tabular}     & \begin{tabular}[c]{@{}c@{}}Intersection over union (mIoU) \\ improved to 0.98 from 0.94 for\\ background class and to 0.89\\ from 0.76 for vegetation.\end{tabular} \\ \hline
\multicolumn{4}{c}{\cite{khan2021novel}}                                                                                                                                                                                                                                                                                                                                                                                                                  \\ \hline
\begin{tabular}[c]{@{}c@{}}Artificial data generated using UAV\\ images for supporting crop/weed\\ species identification at an early stage\end{tabular}  & \begin{tabular}[c]{@{}c@{}}Strawberry \\ and peas\end{tabular} & \begin{tabular}[c]{@{}c@{}}Semi-supervised \\ GAN\\ (SGAN)\end{tabular} & \begin{tabular}[c]{@{}c@{}}Classification accuracy of\\ 90\% was achieved using only\\ 20\% of labelled dataset.\end{tabular}                                       \\ \hline
\multicolumn{4}{c}{Ours}                                                                                                                                                                                                                                                                                                                                                                                                                                                   \\ \hline
\begin{tabular}[c]{@{}c@{}}Data augmentation that uses two GANs \\ to create artificial images to augment the\\ training data\end{tabular}                & Sugarbeet                                                      & \begin{tabular}[c]{@{}c@{}}Shape \\ and Style GAN\end{tabular}          & \begin{tabular}[c]{@{}c@{}}Intersection over union (mIoU) \\ improved to 0.99 from 0.94 for\\ background class and to 0.93\\ from 0.76 for vegetation.\end{tabular} \\ \hline
\end{tabular}

\label{tab:comparison}
\end{table*}

\section{Materials and Methods}
\label{sec:method}

\begin{figure*}[t]
    \centering
    \includegraphics[width=0.99\linewidth]{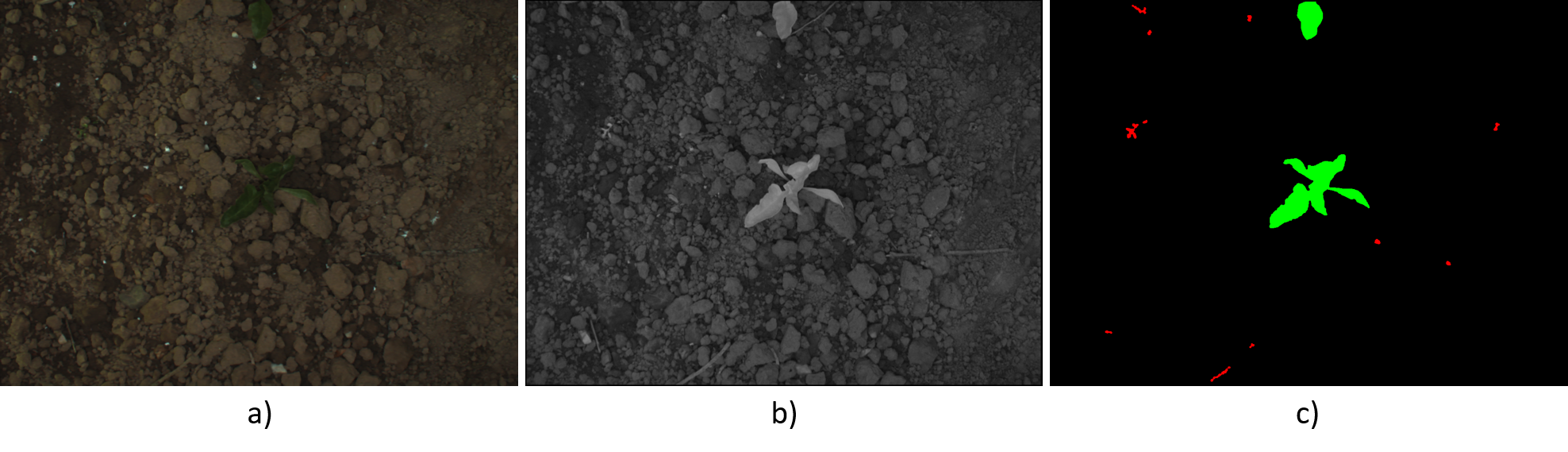}
    \caption{RGB (a), NIR (b), and ground truth (c) samples from the Bonn sugar beet dataset \cite{chebrolu2017ijrr}.}
    \label{fig:bonn}
\end{figure*}

Before describing our strategy, we point out that, in this work, we focus on sugar beets. In particular, we use the publicly available Bonn sugar beet dataset \citep{chebrolu2017ijrr} to demonstrate the effectiveness of our approach. The Bonn sugar beet dataset has been collected through a Bonirob farm robot across different weeks on a sugar beet field. It consists of images captured by a four-channel JAI AD-13 camera (RGB + NIR), mounted on the robot and facing downwards, and annotated at the pixel level. Examples of RGB, NIR, and ground truth images from the Bonn sugar beet dataset are shown in Fig. \ref{fig:bonn}.

\subsection{Proposed Strategy}

\begin{figure}[t]
    \centering
    \includegraphics[width=0.95\columnwidth]{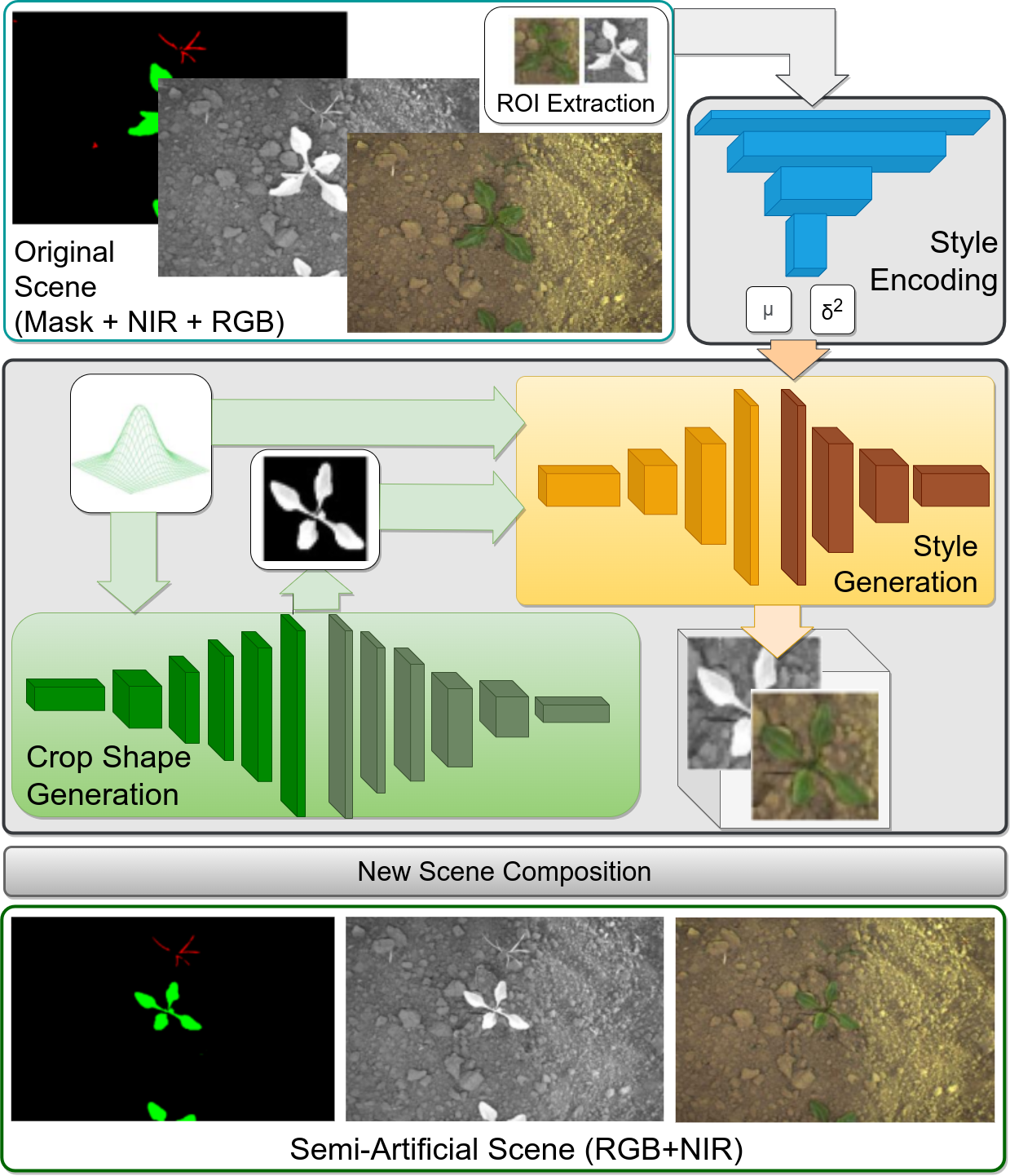}
    \caption{Our pipeline for synthetic shape and style generation. The main input is a real scene (RGB, NIR and ground truth mask) along with Gaussian noise and the final output is Semi-artificial scene.}
    \label{fig:detail shape Style}
\end{figure}

The main objective of our approach is to balance our dataset in order to improve the performance of the crop/weed segmentation task. To achieve our goal, we create semi-artificial images by synthesizing only the crop objects in real images. For the generation process, we consider both the shape and the style of the sugar beet crop. The main steps of the proposed approach are shown in Fig. \ref{fig:detail shape Style}. 

We can summarize the steps of our approach as follows.
\begin{itemize}
    \item Crop shape generation.
    \item Crop and background style generation.
    \item Scene composition (or replacement process).
\end{itemize}

In the first step, we start with generating the new crop shape using a Deep Convolutional Generative Adversarial Network (DCGAN) \citep{DCGAN}, which receives only the normal distribution as input. The shape generator creates small (256 $\times$ 256 pixels) binary patches containing white pixels for the crop and black pixels as background: This shape represents the mask for the synthetic crop. 
In the second step, we start building the RGB image that corresponds to the synthetic mask by generating the texture for the crop through a cGAN. We use as input the mask generated in the previous step and the normal distribution to generate a random style. 
To finish building the RGB image, we must generate the background style (or texture).

An important aspect has to be considered here: The synthetic background should fit with the full image background, so when replacing the real patches with the synthetic ones, we want to preserve the consistency of the background.
For this reason, we encode the original style of the original background by using an image variational autoencoder, getting mean and variance values. Then, we use them along with the generated mask as input to a cGAN to obtain a guided style generation of our background.
The third step concerns building the new semi-artificial image by replacing the original crop patches with the synthetic ones. 

The following sections contain the details for the three above introduced processing steps.

\begin{figure*}[t]
    \centering
    \includegraphics[width=0.99\linewidth]{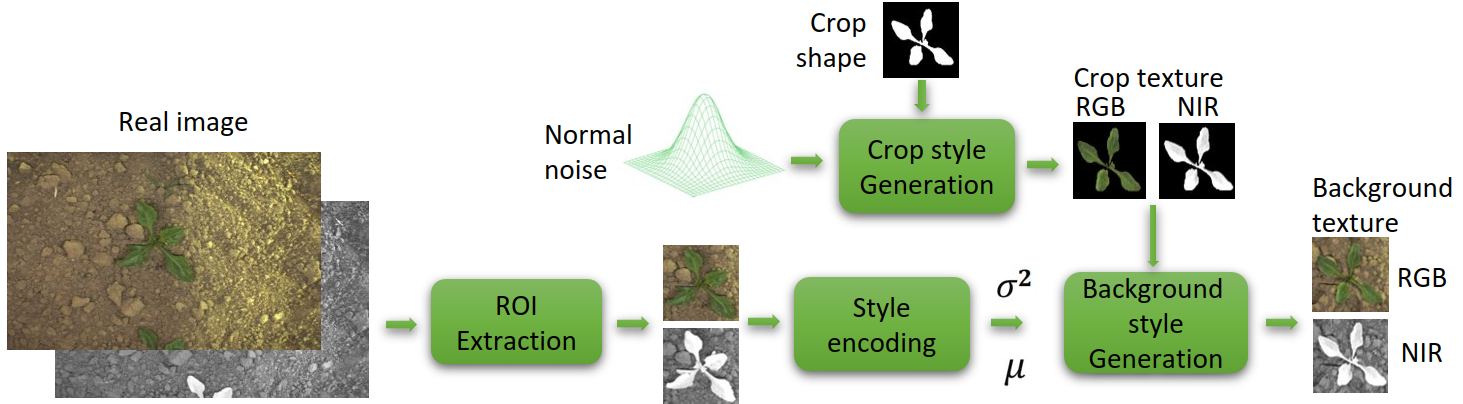}
    \caption{Crop/soil style generation process.}
    \label{fig:style_generation}
\end{figure*}

\subsection{Crop Shape Generation}\label{sec:crop-shape-generation}
The first step in our method concerns the generation of the crop mask in which we use the DCGAN. 
One of the problems to solve in the mask generation process is the difficulty in training a network with images presenting an abrupt change in the border between the crop and the soil. To solve this problem, we used blurred masks during the training process, and this helped the network to learn better how to behave when switching from the crop to the soil.
We also performed several preprocessing steps on the images before feeding them to the DCGAN for training. These
steps included resizing all images to a uniform size, normalizing pixel values to a range of [-1, 1] to stabilize training, applying data augmentation techniques such as rotation, flipping, and cropping to increase dataset diversity.
DCGANs are a direct extension of GANs, thus in the same way as GANs, DCGANs are made of two distinct models, a generator and a discriminator.

\textbf{Generator.}
The generator takes as input the random noise distribution $Z$ with latent size $100$. This layer is a dense layer.
Then, we shape the results into four dimensional tensors and we implement a batch normalization. In particular, we start with a $8\times8$ size. The batch normalization module contains an up-sampling block, which consists of an up-sampling layer followed by a convolutional layer with filter size $3\times3$ and then an activation ReLU layer. This block is repeated for six times to arrive to our target size, which is $256\times256$. After the up-sampling block, we add two convolutional layers and, at the end of the generator, we add an activation layer with $tanh$ as the activation function.

\textbf{Discriminator.}
The main objective of the discriminator is to distinguish between the real and fake generated samples. The first input for the discriminator is the sample coming from the real dataset, which is a small patch of the crop mask
. We represent the crop with white pixels and the background with black pixel.
The second input for the generator is the fake generated sample created in the first stage. The output is a scalar that indicates if the input is coming from the fake or from the real distribution.
The discriminator starts with an input layer of shape $256\times256\times1$, after that we add a Gaussian noise for the samples coming from both the real and fake distributions. According to \cite{arjovsky2017principled}, this makes the results smooth in both data and model probability distributions. After the input layer, we add a convolutional layer, followed by a LeakyReLU activation function and a dropout stage. Then, we have a down-sampling block, which is composed of convolutional layers followed by LeakyReLU and dropout. Finally, we implement a Batch Normalization and we repeat this procedure five times to arrive at the size of $4\times4$. The model ends with flatten and dense layers with a sigmoid activation function. 

\subsection{Crop and Background Style Generation}

After generating the shape of the crop, we need to add some style (i.e., the texture). We start with crop style generation, in which we use the \textit{SPADE} generative adversarial network \citep{spade}. The inputs for our generator are the masks generated in the previous step plus Normal noise. The generator output is the crop with style as shown in Fig. \ref{fig:style_generation}.

When the style of the crop is ready, we generate the style of the background (i.e., the soil). To this end, we use again the SPADE cGAN but, since we need the background style to be aligned with the entire scene, 
we have to guide the generator of the SPADE cGAN. 

To do so, we encode the style of the cropped original background patch from the real images by using a variational autoencoder. Then, we use the encoded style as input to our conditional GAN (SPADE) along with the mask generated in the previous step.

The style encoder is 
composed of a series of convolutional layers with stride 2, followed by two linear layers that output a mean vector $\mu$ and a variance vector $\sigma$. Then, we use $\mu$ and $\sigma$ to compute the noise input to the generator.

For both crop and soil, we train the cGAN with a four-layer image to generate the RGB and NIR images together. This allows the texture of the soil to match both images. 
\subsection{Scene Composition}

\begin{figure}[t]
    \centering
    \includegraphics[width=0.95\columnwidth]{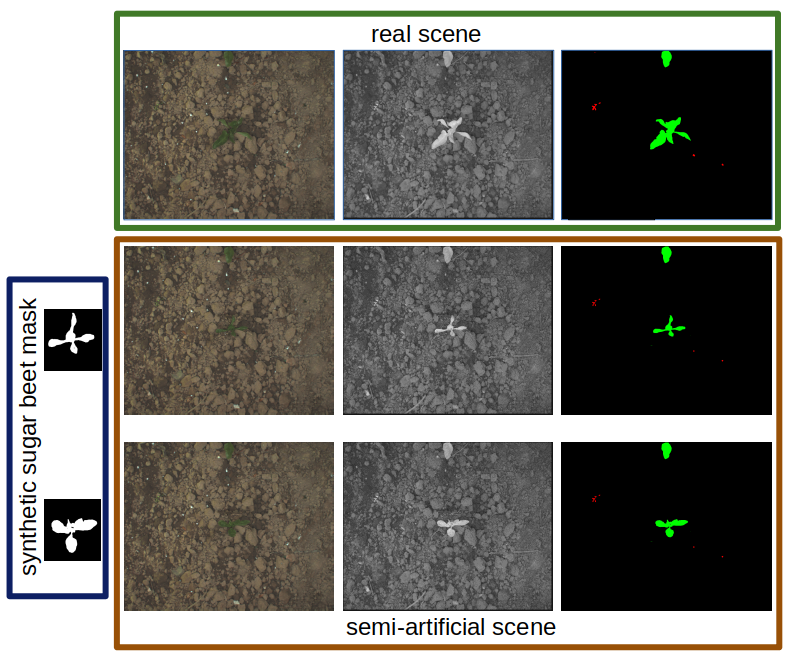}
    \caption{Multispectral synthetic scene generation. (RGB, NIR, and Ground truth). Highlighted in green is the real scene and, on the left highlighted in blue, the new crop shape. The rest of the table represents the synthetic images obtained by inserting in the original image a plant sample generated with our method.
    }
    \label{fig:synthetic_shape_style}
\end{figure}

The final step in our approach concerns the composition of the artificial crop and background with the real scene in order to generate the semi-artificial scene (see Fig. \ref{fig:synthetic_shape_style}).
To this end, we use the ground truth mask of the real scene to extract the real crop patches from the real RGB/NIR image. Then, we replace each crop patch in the real RGB/NIR images with the synthetically generated ones. We consider only the crop objects whose stem is located in the image. 
For the mask replacement, we extract the mask patch that corresponds to the real crop patch that we want to replace. 
For the crop mask patch, we simply replace it with the fake generated crop mask. For the weed mask, we deal with the overlapping between the fake generated crop and the weed by removing the ground truth pixels that belong to the weed mask.

\subsection{Training and Objective Function} \label{sec:trianing_SPADE}
To train the SPADE network we used a learning rate of 0.0001 for the generator and 0.0004 for the discriminator plus the ADAM optimizer with $\beta_1=0$  and $\beta_2 =0.999$ for the generator and discriminator, respectively. The objective function of the SPADE contains the \textbf{Multiscale Adversarial Loss}:

\begin{equation}
\begin{aligned}
    L_D = \mathbb{E}_{(x,y) \sim p_{data}}[min(0,-1+D(x,y))] -\\- \mathbb{E}_{z \sim p_z,y \sim p_{data}}[min(0,-1-D(G(z),y))], \    \\
    L_G = - \mathbb{E}_{z \sim p_z,y \sim p_{data}}0,-1-D(G(z),y)
        \end{aligned}
    \label{eq:multi_loss_eq}
\end{equation}

This loss is implemented in a multiscale way, where we create a pyramid from the generated image by resizing the image to different scales and then, for each scale, we compute the loss. Then, the \textbf{Feature Matching Loss} allows the generator to create images that not only fool the discriminator, but also capture the same statistical properties of the images. To this end, we extract the feature maps from the discriminator for both fake and real images and then compute the $L1$ distance between these two feature maps. This is repeated for all the scales of the generated images:

\begin{equation}
    L_{FM}(G,D_k) = \mathbb{E}_{s,x}   \sum_{i=1}^{T}\dfrac{1}{N_i}[\parallel D_k^{(i)}(s,x)- D_k^{(i)}(s,G(s))\parallel_1]
    \label{eq:Featur_match_lo}
\end{equation}

where $T$ represents the feature maps, $N_i$ is the normalization for each feature map, and $k$ represents the image scale. Finally, the \textbf{VGG loss} is computed as follows.

\begin{equation}
\begin{aligned}
    &L_{VGG}(G,D_k) =
    \mathbb{E}_{s,x} \sum_{i=1}^{5}\dfrac{1}{2^i}[\parallel   VGG(x,M_i)-\\&- VGG(G(s),M_i)\parallel_1]  
    \label{eq:vgg_los_eq}
    \end{aligned}
\end{equation}
where $V GG(x,Mi)$ represents the feature map $M$ of VGG19 and $x$ is the input.

It is worth noticing that, VGG loss is obtained in the same way as the feature matching loss, but with the difference that we compute the feature maps for both real and fake generated by using a VGG19 pre-trained model on \textit{imageNet} dataset, instead of using the discriminator.


We include the encoder in the training process by adding the $KL$ divergence loss:

\begin{equation}
    L_{KLD} = \mathbb{D}_{kl}(q(z\mid x) \parallel p(z))
    \label{eq:kl_eq}
\end{equation}

where $p(z)$ is the standard Gaussian prior distribution $q(z\mid x)$ is the variational distribution, and $q$ is fully determined by a mean and variance vector.
This loss is similar to the loss in the Variational Auto-Encoder \citep{vAE}, where the generator of SPADE GAN plays the role of the decoder.

\begin{figure}[!t]
    \centering
    \includegraphics[width=0.99\columnwidth]{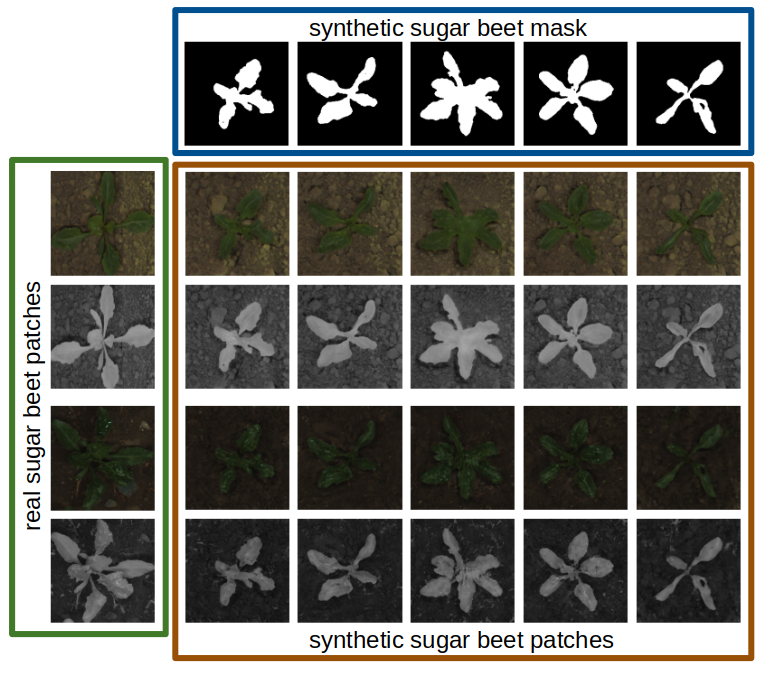}
    \caption{Examples of crop and soil generated with different styles. The first row represents a set of crop/soil masks, while the first column on the left represents the real sugar beet patches used to guide the generation process of the style. The rest represents the synthetic sugar beet patches obtained by using the guided style images and the synthetically generated mask.}
    \label{fig:crop_soil_style}
\end{figure}

Fig. \ref{fig:crop_soil_style} shows some samples generated with different styles. Styles are presented in the column on the left and the masks on the top. From Fig. \ref{fig:crop_soil_style}, it is possible also to visualize that the network has learned how to generate different RGB and NIR styles.

\section{Experimental Results}
\label{sec:results}

We carried out two different experiments, the first to  demonstrate that our method allows to obtain a better
segmentation than traditional approaches and the second to show the contribution of, having both multi-spectral and synthetic data augmentation.

\subsection{Training Generative Adversarial Networks for Shape and Style Generation}
Our DCGAN training process for crop shape generation involves refining both the discriminator and generator networks iteratively. We trained the model using a dataset containing 1000 crop mask patches with size of 256 $\times$ 256 extracted from the Bonn Dataset \cite{chebrolu2017ijrr} at different growth stages of the crop, ensuring a diverse representation. Adversarial ground truths, 'valid' and 'fake', guide the discriminator's classification. The generator synthesizes fake images from random noise, aiming to produce realistic crop shapes. The discriminator learns to distinguish real from generated images, while the generator aims to create indistinguishable crop shapes. 
For training the cGAN we utilize a dataset extracted from the Bonn Dataset, comprising 1000 image patches with the size of 256 $\times$ 256 each for both NIR (Near-Infrared) and RGB channels, alongside corresponding masks as conditional inputs, ensuring a comprehensive representation of various crop and soil conditions.
\subsection{Semantic Segmentation Results}

This experiment aims to show the effectiveness of the proposed approach in improving the mIoU of semantic segmentation. Another objective in this experiment is to compare the proposed approach with traditional augmentation strategies like basic image manipulations (i.e., rotation, shifting, flipping, zooming, and cropping) and texture manipulations (i.e., Gaussian and median blurring, noise injection, contrast, and brightness variation).

We trained Bonnet CNN \citep{milioto2019icra}, with six different datasets, using data from the Bonn sugar beet datasets: 
\begin{enumerate}
    \item \textit{Original}, which is a reduced version of the Bonn dataset. We used a total of 1.600 images, randomly chosen among different days of acquisition in order to contain different growth stages of the target crop. Then, we split it into a training set (1.000 images), a validation set (300 images), and a test set (300 images). 
    \item \textit{Synthetic Crop}, composed of 1.000 images with synthetic crop generated by using our architecture.
    \item \textit{Basic augmentation}: 500 original images augmented with 500 images using basic image operations. 
    \item \textit{Texture augmentation}:500 original images augmented with 500 images using texture manipulations.
    \item \textit{Style augmentation}: composed by the union of 500 images from the Original dataset and 500 images with synthetic crop generated by the method in \cite{FAWAKHERJI2021103861}.
    \item \textit{Style and Shape augmentation}, containing 1.000 images: 500 images from the Original dataset and 500 synthetic crop images generated by the proposed approach.
\end{enumerate}

For the synthetic datasets, we have replaced with synthetic samples only those plants whose stems are fully framed in the image.
For the plants that are mostly out of the frame, the original one is kept. We experimentally verified that it is necessary to have the stem of the plant roughly in the center the of mask, to obtain an effective synthetic image generation.

To evaluate the semantic segmentation output, we used the Mean Intersection over Union (denoted as \textit{mIoU}).
Quantitative results of the semantic segmentation on real images from the Bonn dataset are shown in Table~\ref{tab:augmentation_shape}. The results prove that the IoU increases by using the original dataset augmented with the synthetic ones compared to using only the original dataset. 
Additionally, the rate of correctly predicted crop and weed samples increases when we use the mixed dataset for training. The correctly predicted samples increase more than $19\%$ in the case of sugar beet, and around $6\%$ for weed samples w.r.t. the \textit{Original} dataset.
Moreover, using only the synthetic dataset also leads to a competitive performance when compared to using only the original one. 


\begin{table}[t]
\centering
\caption{Segmentation results of Bonnet architecture, trained on six different datasets, tested on Bonn test dataset}
\begin{tabular}{l c c c }
\cline{1-4}
  \multirow{2}{*}{Augmentation Strategy} &  &  \multicolumn{2}{c}{IoU}   \\ 
 \cline{3-4}
& mIoU & Crop  &
     Weed \\ 
\hline
\textit{Original} & 0.70 &  0.75 & 0.35\\ \textit{Synthetic Crop}     &0.69  & 0.74  & 0.34\\
\textit{Basic augmentation} & 0.71 & 0.76 & 0.37 \\ 
\textit{Texture augmentation} & 0.73 & 0.79 & 0.40 \\ 
\textit{Style augmentation} & 0.76  & 0.92 & 0.37
      \\   
\textit{Shape and Style augmentation} & \textbf{0.78}    &  \textbf{0.94}  & \textbf{0.41}    \\   
\hline
\end{tabular}
\label{tab:augmentation_shape}
\end{table}


\subsection{Synthetic Multi-Spectral Images Evaluation}

  To show the contribution of having both multi-spectral and synthetic data augmentation, we considered four different training sets, i.e., \textit{Original} and \textit{Mixed} containing RGB images only and \textit{Original} and \textit{Mixed} containing both RGB and NIR images. Table~\ref{tab:Multi_channel_shape_results} shows the segmentation results for this experiment. The segmentation capability improves when using the \textit{Mixed} dataset, i.e., when the dataset containing real images is augmented with synthetic data. This supports the idea of creating artificial samples to improve the segmentation performance.

Furthermore, the results in Table~\ref{tab:Multi_channel_shape_results} show that using the \textit{Mixed} RGB plus NIR dataset during the training process leads to a better performance. 
This supports our claim that also the NIR channel generated using our approach improves the segmentation capability of the convolutional network architecture used in our experiments.

\begin{table}[t]
\caption{Pixel-wise segmentation performance, networks trained on two different inputs (RGB and RGB + NIR), tested on Bonn test dataset.}
\begin{center}
\begin{tabular}{ c  c   c   c  }
 \hline
 Train set &  & \multicolumn{2}{c}{IOU} \\ \cline{3-4} 
                          & mIoU & Crop & Weed    \\   \hline

\textit{Original (RGB)}  
                          & 0.70 & 0.75 & 0.35     \\ \hline
\textit{Mixed (RGB) } 
                           & 0.78 & 0.94  & 0.41\\ \hline
\textit{Original (RGB+NIR)} 
                           & 0.77 & 0.85 & 0.45        \\ \hline
\textit{Mixed (RGB+NIR)}    
                          & \textbf{0.82} & \textbf{0.95} & \textbf{0.53}        \\  \hline

\end{tabular}
\end{center}
\label{tab:Multi_channel_shape_results}
\end{table}

\subsection{Ablation Test}
As a further demonstration of the validity of our approach, we extend the experiments by focusing only on 256 $\times$ 256 patches representing a single instance of the crop. We have performed the training on the following three datasets:
\begin{itemize}
    \item \emph{Real}: 2.000 crop and soil real patches extracted from the Bonn dataset.
    \item \emph{Real + style augmentation}: \emph{Real} dataset augmented with 500 patches generated by style GAN.
    \item \emph{Real + shape and style augmentation}: \emph{Original} dataset augmented with 500 patches generated by the proposed approach.
\end{itemize}
For testing, we used 400 images of sugar beet patches extracted from the Bonn sugar beet dataset not used in the training phase.
The comparison results, presented in Table \ref{tab:Patch_evaluation}, show that the model trained with the data augmented with shape and style overcomes both the models trained with the real data and data augmented with style only.

\begin{table}[t]
\centering
\caption{Pixel-wise segmentation performance for Bonnet architecture, trained on three different datasets.}
\begin{tabular}{l c c c }
\hline
  \multirow{2}{*}{Model} &  &  \multicolumn{2}{c}{IoU}   \\ 
 \cline{3-4}
& mIoU & Soil  &
     Crop \\
\hline
\textit{Real} & 0.85 & 0.94 & 0.76 \\ 
\textit{Real + style augmentation} &  0.94&0.98&
     0.89   \\   
\textit{Real + shape and style augmentation}& \textbf{0.96} & \textbf{0.99}& \textbf{0.93} \\ 
\hline
\end{tabular}
\label{tab:Patch_evaluation}
\end{table}

\section{Conclusions}
\label{sec:conclusions}
In this paper, we have presented a data augmentation strategy for improving segmentation that exploits two types of GANs, namely DCGAN and cGAN, to generate entire agricultural scenes by synthesizing only the most relevant objects. 
The core of the proposed approach lies in exploiting the shapes of real objects to condition the trained generative models. The existing shapes are extracted from real-world labeled images. In addition, the generation process also synthesizes the NIR channel. The synthetically augmented dataset, obtained in this way, can then be used to train a semantic segmentation network.

We introduced a shape and style augmentation approach, in which we augment the style and the shape of the target object: To generate the shape, we used a DCGAN and then, we used the first approach to build the style of the target object.
We applied our method to the crop/weed segmentation problem.

Different kinds of quantitative evaluation have been carried out to demonstrate that augmenting datasets with our approach can improve the performance of state-of-the-art segmentation architectures. The experimental results show that the segmentation quality increases by using the real dataset augmented with synthetic data.   

\section*{Acknowledgement}
\euflag \quad This work is part of a project that has received funding from the European Union’s Horizon 2020 research and innovation programme under grant agreement No 101016906 – Project CANOPIES

\euflag \quad This work has been partially supported by project AGRITECH Spoke 9 - Codice progetto MUR: AGRITECH ”National Research Centre for Agricultural Technologies” - CUP CN00000022, of the National Recovery and Resilience Plan (PNRR) financed by the European Union ”Next Generation EU”.

\section*{CRediT authorship contribution statement 
}

Mulham Fawakherji: Writing – original draft, Methodology, Investigation, Formal analysis, Conceptualization.

Vincenzo Suriani: Writing, Formal analysis, Testing methodology, Data curation.

Daniele Nardi: Project administration, Funding acquisition, Conceptualization, Review \& Editing.

Domenico Daniele Bloisi: Resources, Supervision, Conceptualization, Review \& Editing.
\bibliographystyle{elsarticle-harv}

\bibliography{cas-refs}

\begin{thebibliography}{26}
\expandafter\ifx\csname natexlab\endcsname\relax\def\natexlab#1{#1}\fi
\providecommand{\url}[1]{\texttt{#1}}
\providecommand{\href}[2]{#2}
\providecommand{\path}[1]{#1}
\providecommand{\DOIprefix}{doi:}
\providecommand{\ArXivprefix}{arXiv:}
\providecommand{\URLprefix}{URL: }
\providecommand{\Pubmedprefix}{pmid:}
\providecommand{\doi}[1]{\href{http://dx.doi.org/#1}{\path{#1}}}
\providecommand{\Pubmed}[1]{\href{pmid:#1}{\path{#1}}}
\providecommand{\bibinfo}[2]{#2}
\ifx\xfnm\relax \def\xfnm[#1]{\unskip,\space#1}\fi
\bibitem[{Arjovsky and Bottou(2017)}]{arjovsky2017principled}
\bibinfo{author}{Arjovsky, M.}, \bibinfo{author}{Bottou, L.},
  \bibinfo{year}{2017}.
\newblock \bibinfo{title}{Towards principled methods for training generative
  adversarial networks}.
\newblock \href{http://arxiv.org/abs/1701.04862}{{\tt arXiv:1701.04862}}.
\bibitem[{Badrinarayanan et~al.(2015)Badrinarayanan, Kendall and
  Cipolla}]{Segnet}
\bibinfo{author}{Badrinarayanan, V.}, \bibinfo{author}{Kendall, A.},
  \bibinfo{author}{Cipolla, R.}, \bibinfo{year}{2015}.
\newblock \bibinfo{title}{Segnet: {A} deep convolutional encoder-decoder
  architecture for image segmentation}.
\newblock \bibinfo{journal}{CoRR} \bibinfo{volume}{abs/1511.00561}.
\newblock \href{http://arxiv.org/abs/1511.00561}{{\tt arXiv:1511.00561}}.
\bibitem[{Chebrolu et~al.(2017)Chebrolu, Lottes, Schaefer, Winterhalter,
  Burgard and Stachniss}]{chebrolu2017ijrr}
\bibinfo{author}{Chebrolu, N.}, \bibinfo{author}{Lottes, P.},
  \bibinfo{author}{Schaefer, A.}, \bibinfo{author}{Winterhalter, W.},
  \bibinfo{author}{Burgard, W.}, \bibinfo{author}{Stachniss, C.},
  \bibinfo{year}{2017}.
\newblock \bibinfo{title}{Agricultural robot dataset for plant classification,
  localization and mapping on sugar beet fields}.
\newblock \bibinfo{journal}{The International Journal of Robotics Research} .
\bibitem[{Di~Cicco et~al.(2017)Di~Cicco, Potena, Grisetti and
  Pretto}]{dicicco2017}
\bibinfo{author}{Di~Cicco, M.}, \bibinfo{author}{Potena, C.},
  \bibinfo{author}{Grisetti, G.}, \bibinfo{author}{Pretto, A.},
  \bibinfo{year}{2017}.
\newblock \bibinfo{title}{Automatic model based dataset generation for fast and
  accurate crop and weeds detection}, in: \bibinfo{booktitle}{IROS},
  \bibinfo{organization}{IEEE}. pp. \bibinfo{pages}{5188--5195}.
\bibitem[{Divyanth et~al.(2022)Divyanth, Guru, Soni, Machavaram, Nadimi and
  Paliwal}]{divyanth2022image}
\bibinfo{author}{Divyanth, L.}, \bibinfo{author}{Guru, D.},
  \bibinfo{author}{Soni, P.}, \bibinfo{author}{Machavaram, R.},
  \bibinfo{author}{Nadimi, M.}, \bibinfo{author}{Paliwal, J.},
  \bibinfo{year}{2022}.
\newblock \bibinfo{title}{Image-to-image translation-based data augmentation
  for improving crop/weed classification models for precision agriculture
  applications}.
\newblock \bibinfo{journal}{Algorithms} \bibinfo{volume}{15},
  \bibinfo{pages}{401}.
\bibitem[{Espejo-Garcia et~al.(2021)Espejo-Garcia, Mylonas, Athanasakos, Vali
  and Fountas}]{espejo2021combining}
\bibinfo{author}{Espejo-Garcia, B.}, \bibinfo{author}{Mylonas, N.},
  \bibinfo{author}{Athanasakos, L.}, \bibinfo{author}{Vali, E.},
  \bibinfo{author}{Fountas, S.}, \bibinfo{year}{2021}.
\newblock \bibinfo{title}{Combining generative adversarial networks and
  agricultural transfer learning for weeds identification}.
\newblock \bibinfo{journal}{Biosystems Engineering} \bibinfo{volume}{204},
  \bibinfo{pages}{79--89}.
\bibitem[{Fawakherji et~al.(2021)Fawakherji, Potena, Pretto, Bloisi and
  Nardi}]{FAWAKHERJI2021103861}
\bibinfo{author}{Fawakherji, M.}, \bibinfo{author}{Potena, C.},
  \bibinfo{author}{Pretto, A.}, \bibinfo{author}{Bloisi, D.D.},
  \bibinfo{author}{Nardi, D.}, \bibinfo{year}{2021}.
\newblock \bibinfo{title}{Multi-spectral image synthesis for crop/weed
  segmentation in precision farming}.
\newblock \bibinfo{journal}{Robotics and Autonomous Systems}
  \bibinfo{volume}{146}, \bibinfo{pages}{103861}.
\newblock \DOIprefix\doi{https://doi.org/10.1016/j.robot.2021.103861}.
\bibitem[{Fawakherji et~al.(2019)Fawakherji, Youssef, Bloisi, Pretto and
  Nardi}]{FawakherjiYBPN19}
\bibinfo{author}{Fawakherji, M.}, \bibinfo{author}{Youssef, A.},
  \bibinfo{author}{Bloisi, D.}, \bibinfo{author}{Pretto, A.},
  \bibinfo{author}{Nardi, D.}, \bibinfo{year}{2019}.
\newblock \bibinfo{title}{Crop and weeds classification for precision
  agriculture using context-independent pixel-wise segmentation}, in:
  \bibinfo{booktitle}{IRC}, pp. \bibinfo{pages}{146--152}.
\bibitem[{Giuffrida et~al.(2017)Giuffrida, Scharr and
  Tsaftaris}]{valerio2017arigan}
\bibinfo{author}{Giuffrida, M.V.}, \bibinfo{author}{Scharr, H.},
  \bibinfo{author}{Tsaftaris, S.A.}, \bibinfo{year}{2017}.
\newblock \bibinfo{title}{{ARIGAN}: Synthetic arabidopsis plants using
  generative adversarial network}, in: \bibinfo{booktitle}{Proceedings of the
  IEEE International Conference on Computer Vision Workshops}, pp.
  \bibinfo{pages}{2064--2071}.
\bibitem[{Khan et~al.(2021)Khan, Tufail, Khan, Khan, Iqbal and
  Alam}]{khan2021novel}
\bibinfo{author}{Khan, S.}, \bibinfo{author}{Tufail, M.},
  \bibinfo{author}{Khan, M.T.}, \bibinfo{author}{Khan, Z.A.},
  \bibinfo{author}{Iqbal, J.}, \bibinfo{author}{Alam, M.},
  \bibinfo{year}{2021}.
\newblock \bibinfo{title}{A novel semi-supervised framework for uav based
  crop/weed classification}.
\newblock \bibinfo{journal}{Plos one} \bibinfo{volume}{16},
  \bibinfo{pages}{e0251008}.
\bibitem[{Kim and Park(2022)}]{KIM2022107146}
\bibinfo{author}{Kim, Y.H.}, \bibinfo{author}{Park, K.R.},
  \bibinfo{year}{2022}.
\newblock \bibinfo{title}{Mts-cnn: Multi-task semantic
  segmentation-convolutional neural network for detecting crops and weeds}.
\newblock \bibinfo{journal}{Computers and Electronics in Agriculture}
  \bibinfo{volume}{199}, \bibinfo{pages}{107146}.
\newblock \URLprefix
  \url{https://www.sciencedirect.com/science/article/pii/S016816992200463X},
  \DOIprefix\doi{https://doi.org/10.1016/j.compag.2022.107146}.
\bibitem[{Kingma and Welling(2014)}]{vAE}
\bibinfo{author}{Kingma, D.P.}, \bibinfo{author}{Welling, M.},
  \bibinfo{year}{2014}.
\newblock \bibinfo{title}{Auto-encoding variational bayes}.
\newblock \href{http://arxiv.org/abs/1312.6114}{{\tt arXiv:1312.6114}}.
\bibitem[{Lottes et~al.(2017)Lottes, Hörferlin, Sander and
  Stachniss}]{lottesJFR2016}
\bibinfo{author}{Lottes, P.}, \bibinfo{author}{Hörferlin, M.},
  \bibinfo{author}{Sander, S.}, \bibinfo{author}{Stachniss, C.},
  \bibinfo{year}{2017}.
\newblock \bibinfo{title}{Effective vision-based classification for separating
  sugar beets and weeds for precision farming}.
\newblock \bibinfo{journal}{Journal of Field Robotics} \bibinfo{volume}{34},
  \bibinfo{pages}{1160--1178}.
\bibitem[{Lu et~al.(2022)Lu, Chen, Olaniyi and Huang}]{LU2022107208}
\bibinfo{author}{Lu, Y.}, \bibinfo{author}{Chen, D.}, \bibinfo{author}{Olaniyi,
  E.}, \bibinfo{author}{Huang, Y.}, \bibinfo{year}{2022}.
\newblock \bibinfo{title}{Generative adversarial networks (gans) for image
  augmentation in agriculture: A systematic review}.
\newblock \bibinfo{journal}{Computers and Electronics in Agriculture}
  \bibinfo{volume}{200}, \bibinfo{pages}{107208}.
\newblock \URLprefix
  \url{https://www.sciencedirect.com/science/article/pii/S0168169922005233},
  \DOIprefix\doi{https://doi.org/10.1016/j.compag.2022.107208}.
\bibitem[{Madsen et~al.(2019)Madsen, Dyrmann, Jørgensen and
  Karstoft}]{MADSEN2019147}
\bibinfo{author}{Madsen, S.L.}, \bibinfo{author}{Dyrmann, M.},
  \bibinfo{author}{Jørgensen, R.N.}, \bibinfo{author}{Karstoft, H.},
  \bibinfo{year}{2019}.
\newblock \bibinfo{title}{Generating artificial images of plant seedlings using
  generative adversarial networks}.
\newblock \bibinfo{journal}{Biosystems Engineering} \bibinfo{volume}{187},
  \bibinfo{pages}{147 -- 159}.
\newblock \DOIprefix\doi{https://doi.org/10.1016/j.biosystemseng.2019.09.005}.
\bibitem[{McCool et~al.(2017)McCool, Perez and Upcroft}]{mccool2017mixtures}
\bibinfo{author}{McCool, C.}, \bibinfo{author}{Perez, T.},
  \bibinfo{author}{Upcroft, B.}, \bibinfo{year}{2017}.
\newblock \bibinfo{title}{Mixtures of lightweight deep convolutional neural
  networks: Applied to agricultural robotics}.
\newblock \bibinfo{journal}{IEEE Robotics and Automation Letters}
  \bibinfo{volume}{2}, \bibinfo{pages}{1344--1351}.
\bibitem[{{Milioto} et~al.(2018){Milioto}, {Lottes} and
  {Stachniss}}]{miliotoicra2018}
\bibinfo{author}{{Milioto}, A.}, \bibinfo{author}{{Lottes}, P.},
  \bibinfo{author}{{Stachniss}, C.}, \bibinfo{year}{2018}.
\newblock \bibinfo{title}{Real-time semantic segmentation of crop and weed for
  precision agriculture robots leveraging background knowledge in cnns}, in:
  \bibinfo{booktitle}{ICRA}, pp. \bibinfo{pages}{2229--2235}.
\bibitem[{Milioto and Stachniss(2019)}]{milioto2019icra}
\bibinfo{author}{Milioto, A.}, \bibinfo{author}{Stachniss, C.},
  \bibinfo{year}{2019}.
\newblock \bibinfo{title}{{Bonnet: An Open-Source Training and Deployment
  Framework for Semantic Segmentation in Robotics using CNNs}}, in:
  \bibinfo{booktitle}{ICRA}.
\bibitem[{{Nguyen Thanh Le} et~al.(2019){Nguyen Thanh Le}, Apopei and
  Alameh}]{NGUYENTHANHLE2019116}
\bibinfo{author}{{Nguyen Thanh Le}, V.}, \bibinfo{author}{Apopei, B.},
  \bibinfo{author}{Alameh, K.}, \bibinfo{year}{2019}.
\newblock \bibinfo{title}{Effective plant discrimination based on the
  combination of local binary pattern operators and multiclass support vector
  machine methods}.
\newblock \bibinfo{journal}{Information Processing in Agriculture}
  \bibinfo{volume}{6}, \bibinfo{pages}{116--131}.
\newblock \DOIprefix\doi{https://doi.org/10.1016/j.inpa.2018.08.002}.
\bibitem[{Park et~al.(2019)Park, Liu, Wang and Zhu}]{spade}
\bibinfo{author}{Park, T.}, \bibinfo{author}{Liu, M.Y.}, \bibinfo{author}{Wang,
  T.C.}, \bibinfo{author}{Zhu, J.Y.}, \bibinfo{year}{2019}.
\newblock \bibinfo{title}{Semantic image synthesis with spatially-adaptive
  normalization}, in: \bibinfo{booktitle}{Proceedings of the IEEE Conference on
  Computer Vision and Pattern Recognition}, pp. \bibinfo{pages}{2337--2346}.
\bibitem[{Potena et~al.(2016)Potena, Nardi and Pretto}]{potena2016}
\bibinfo{author}{Potena, C.}, \bibinfo{author}{Nardi, D.},
  \bibinfo{author}{Pretto, A.}, \bibinfo{year}{2016}.
\newblock \bibinfo{title}{Fast and accurate crop and weed identification with
  summarized train sets for precision agriculture}, in:
  \bibinfo{booktitle}{IAS}, pp. \bibinfo{pages}{105--121}.
\bibitem[{Radford et~al.(2016)Radford, Metz and Chintala}]{DCGAN}
\bibinfo{author}{Radford, A.}, \bibinfo{author}{Metz, L.},
  \bibinfo{author}{Chintala, S.}, \bibinfo{year}{2016}.
\newblock \bibinfo{title}{Unsupervised representation learning with deep
  convolutional generative adversarial networks}.
\newblock \href{http://arxiv.org/abs/1511.06434}{{\tt arXiv:1511.06434}}.
\bibitem[{Sa et~al.(2017)Sa, Chen, Popovic, Khanna, Liebisch, Nieto and
  Siegwart}]{weednet}
\bibinfo{author}{Sa, I.}, \bibinfo{author}{Chen, Z.}, \bibinfo{author}{Popovic,
  M.}, \bibinfo{author}{Khanna, R.}, \bibinfo{author}{Liebisch, F.},
  \bibinfo{author}{Nieto, J.}, \bibinfo{author}{Siegwart, R.},
  \bibinfo{year}{2017}.
\newblock \bibinfo{title}{Weednet: Dense semantic weed classification using
  multispectral images and mav for smart farming}.
\newblock \bibinfo{journal}{IEEE Robotics and Automation Letters}
  \bibinfo{volume}{PP}.
\newblock \DOIprefix\doi{10.1109/LRA.2017.2774979}.
\bibitem[{Wang and Yao(2012)}]{Wang6170916}
\bibinfo{author}{Wang, S.}, \bibinfo{author}{Yao, X.}, \bibinfo{year}{2012}.
\newblock \bibinfo{title}{Multiclass imbalance problems: Analysis and potential
  solutions}.
\newblock \bibinfo{journal}{IEEE Transactions on Systems, Man, and Cybernetics,
  Part B (Cybernetics)} \bibinfo{volume}{42}, \bibinfo{pages}{1119--1130}.
\newblock \DOIprefix\doi{10.1109/TSMCB.2012.2187280}.
\bibitem[{Xu et~al.(2022)Xu, Yoon, Fuentes, Yang and Park}]{xu2022style}
\bibinfo{author}{Xu, M.}, \bibinfo{author}{Yoon, S.}, \bibinfo{author}{Fuentes,
  A.}, \bibinfo{author}{Yang, J.}, \bibinfo{author}{Park, D.S.},
  \bibinfo{year}{2022}.
\newblock \bibinfo{title}{Style-consistent image translation: A novel data
  augmentation paradigm to improve plant disease recognition}.
\newblock \bibinfo{journal}{Frontiers in Plant Science} \bibinfo{volume}{12},
  \bibinfo{pages}{773142}.
\bibitem[{Zhang et~al.(2019)Zhang, Goodfellow, Metaxas and
  Odena}]{zhang2019selfattention}
\bibinfo{author}{Zhang, H.}, \bibinfo{author}{Goodfellow, I.},
  \bibinfo{author}{Metaxas, D.}, \bibinfo{author}{Odena, A.},
  \bibinfo{year}{2019}.
\newblock \bibinfo{title}{Self-attention generative adversarial networks}.
\newblock \href{http://arxiv.org/abs/1805.08318}{{\tt arXiv:1805.08318}}.

\end{thebibliography}



\end{document}